\documentclass[10pt,conference]{IEEEtran}
\makeatletter
\IEEEoverridecommandlockouts
\usepackage{xspace}
\usepackage{graphicx}
\usepackage{subfig}
\usepackage{multicol}
\usepackage{graphicx}
\usepackage{xcolor,colortbl}
\usepackage{gensymb}
\usepackage{amsmath}
\usepackage{hyperref}
\usepackage{verbatim}
\usepackage{textcomp}
\usepackage{amsmath}
\usepackage[vlined,ruled]{algorithm2e}
\usepackage{color}
\usepackage{multirow}
\usepackage{float}
\usepackage{balance}
\usepackage{makecell}
\usepackage{tabularx}
\usepackage{float}
\usepackage{footnote}

\newcommand\Mark[1]{\textsuperscript#1}

\begin{document}

\title{Let me join you! Real-time F-formation recognition \\by a socially aware robot}

% % author names and affiliations
% % use a multiple column layout for up to three different
\author{\IEEEauthorblockN{Hrishav Bakul Barua\Mark{1}, Pradip Pramanick\Mark{1}, Chayan Sarkar\Mark{1}, and Theint Haythi Mg\Mark{2}}
\IEEEauthorblockA{\Mark{1}TCS Research \& Innovation, India\\
\Mark{2}Myanmar Institute of Information Technology, Mandalay, Myanmar}
}

\maketitle
\thispagestyle{empty}
\pagestyle{empty}

%%%%%%%%%%%%%%%%%%%%%%%%%%%%%%%%%%%%%%%%%%%%%%%%%%%%%%%%%%%%%%%%%%%%%%%%%%%%%%%%
\begin{abstract}
This paper presents a novel architecture to detect social groups in real-time from a continuous image stream of an ego-vision camera. F-formation defines social orientations in space where two or more person tends to communicate in a social place. Thus, essentially, we detect F-formations in social gatherings such as meetings, discussions, etc. and predict the robot's approach angle if it wants to join the social group. Additionally, we also detect outliers, i.e., the persons who are not part of the group under consideration. Our proposed pipeline consists of -- a) a skeletal key points estimator (a total of 17) for the detected human in the scene, b) a learning model (using a feature vector based on the skeletal points) using CRF to detect groups of people and outlier person in a scene, and c) a separate learning model using a multi-class Support Vector Machine (SVM) to predict the exact F-formation of the group of people in the current scene and the angle of approach for the viewing robot. The system is evaluated using two data-sets. The results show that the group and outlier detection in a scene using our method establishes an accuracy of \textbf{91\%}. We have made rigorous comparisons of our systems with a state-of-the-art F-formation detection system and found that it outperforms the state-of-the-art by \textbf{29\%} for formation detection and \textbf{55\%} for combined detection of the formation and approach angle.      
\end{abstract}

\vspace{0.25cm}
%\section*{}

%%%%%%%%%%%%%%%%%%%%%%%%%%%%%%%%%%%%%%%%%%%%%%%%%%%%%%%%%%%%%%%%%%%%%%%%%%%%%%%%
\section{Introduction}
\label{sec:intro}
Social robotics has gained exponential momentum in recent years, where the aim is to make the robot behave acceptably and safely in social setups. Some of the interesting use cases of popular robotics applications are telepresence robots \cite{bakul2020can,sau2020demo}, teleoperation robots, service robots, co-worker robots~\cite{pramanick2018defatigue}, etc. In many such applications, robots often need to join a group of people for interaction~\cite{pathi2019f}. People in a group tend to maintain a pattern while they interact with each other. These patterns usually adhere to some orientations and distances among the participating people. So, the robots should be aware of the societal norms of joining existing groups for meetings and discussions  \cite{kruse2013human}, \cite{charalampous2017recent}.

\begin{figure}
	\centering
	\includegraphics[width=\linewidth]{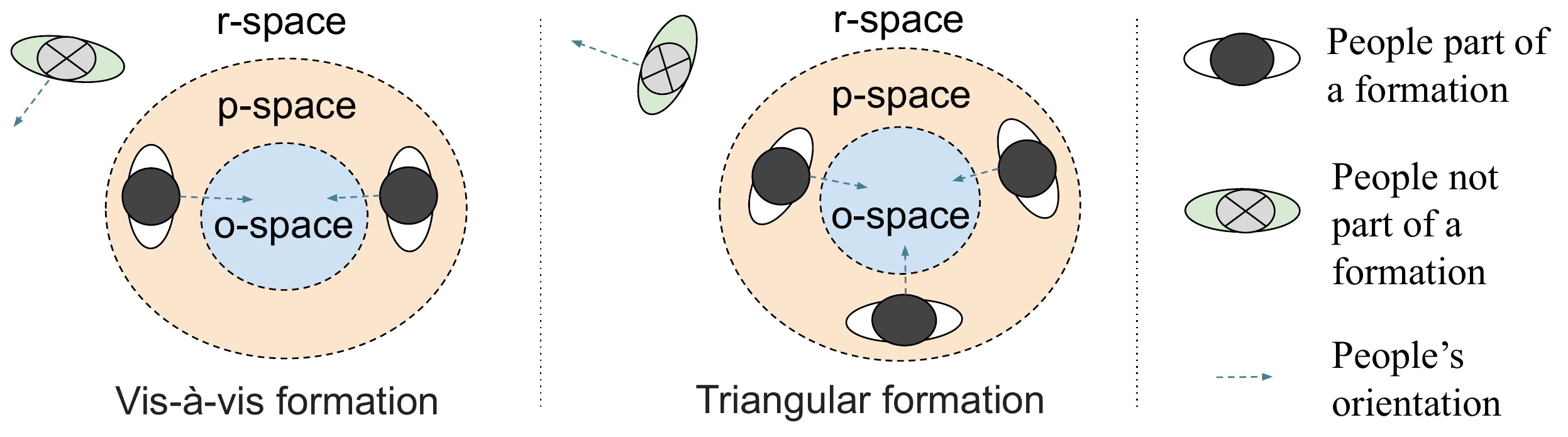}
	\caption{Face-to-face and Triangular F-formation with the o-space, p-space and r-space marked.}
	\label{fig:f_formation_spaces}
\end{figure}

F-formation is a theory proposed by Kendon~\cite{kendon1990conducting} that has been widely realized in group formations among people. It generally consists of three major areas or spaces, which define the intimacy level and societal norms of the interaction. Fig.~\ref{fig:f_formation_spaces} depicts a typical face-to-face (or vis-a-vis) and triangular (or triangle) F-formation. As described by Kendon, p-space is the area where the people stand in a patter, o-space is where the standing people look towards, generally inwards towards each other, and the r-space is the area that lies outside them. Generally, people in r-space are not considered as a part of the conversing group.

\textbf{Motivation}. The primary focus of this work is to make robots respect the societal norms while joining a group like a fellow human being. This will lead to more acceptability of a human towards a robot in social setups and would help a person in developing a certain level of confidence towards a robot. The existing methods to detect F-formations are mostly rule-based. The accuracy of such systems in a dynamic environment is not up to the mark. Moreover, if the group is being viewed from such an angle that some parts of the group are occluded, then the existing methods do not perform acceptably. Secondly, the existing works estimate F-formation based on the head pose and orientation, which may result in the wrong detection as someone might change his/her head orientation for a second and then again get back to his/her original orientation without changing the body orientation. So, the temporal information should also be taken into account in such dynamic cases. Such a method also finds application in \textit{COVID-19} scenario for monitoring human social distancing for ensuring safety and well being of the people in this difficult time.

\begin{figure}
	\centering
	\includegraphics[width=\linewidth]{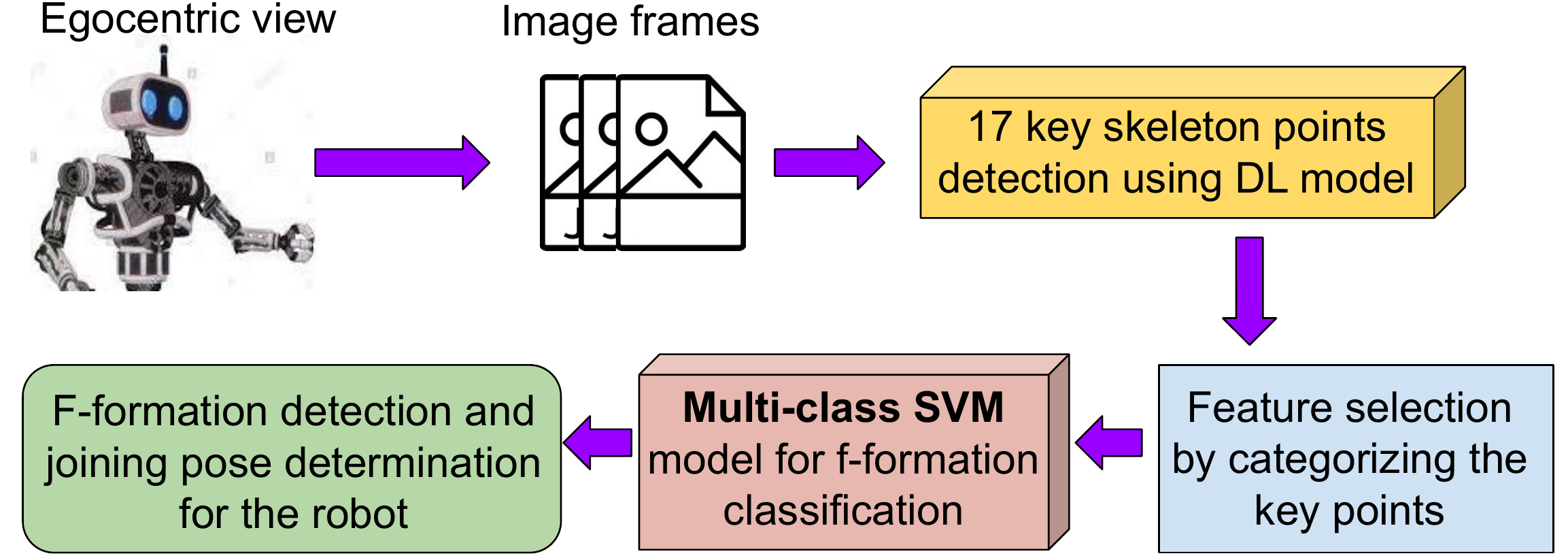}
	\caption{A high-level pipeline to detect F-formation and classify it from live camera feed in a robot.}
	\label{fig:pipeline}
\end{figure}

\begin{figure}
	\centering
	\subfloat[triangle -90]{\includegraphics[width=0.20\textwidth]{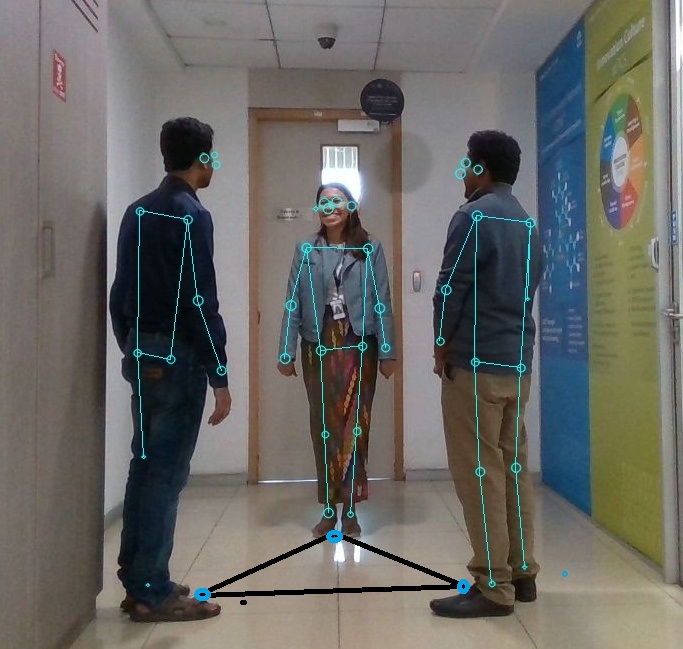}\label{fig:triangle0}}
	\subfloat[L-shaped -90]{\includegraphics[width=0.19\textwidth]{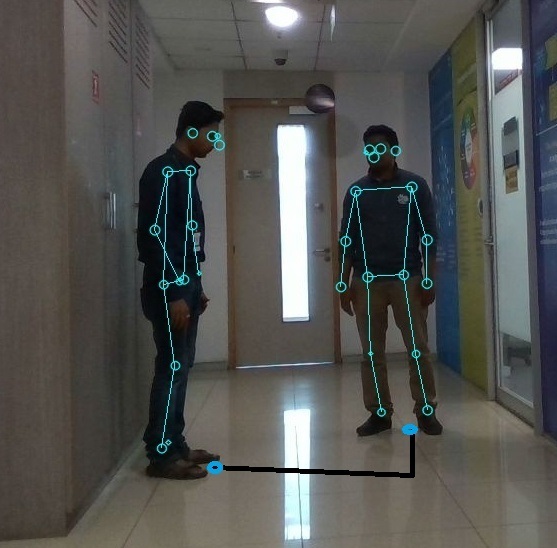}\label{fig:lshaped0}}\\
	\subfloat[side-by-side -90]{\includegraphics[width=0.20\textwidth]{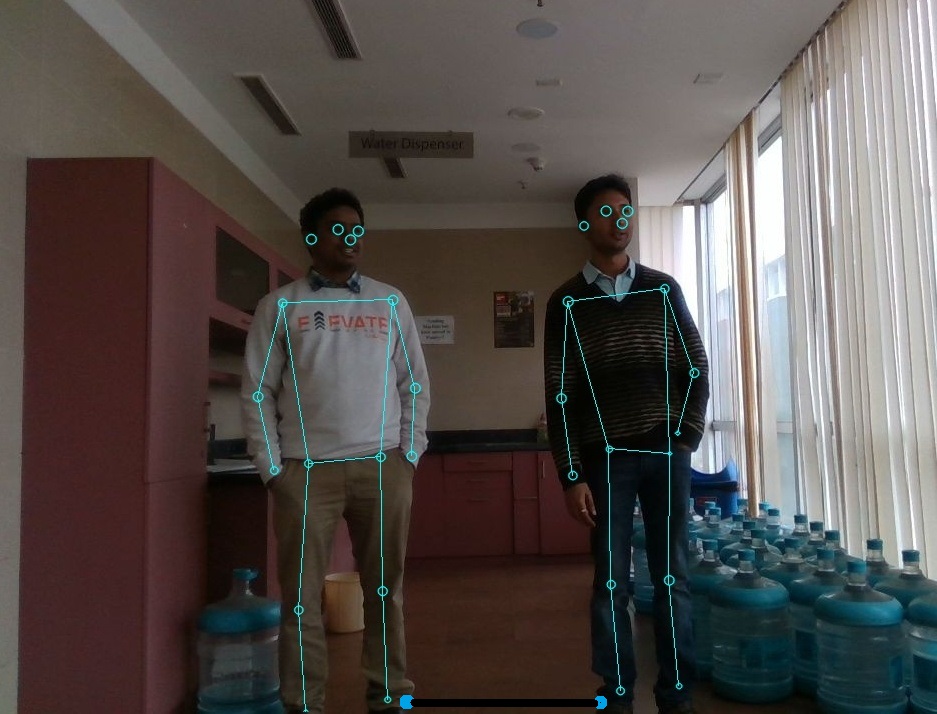}\label{fig:side0}}
	\subfloat[face-to-face -90]{\includegraphics[width=0.19\textwidth]{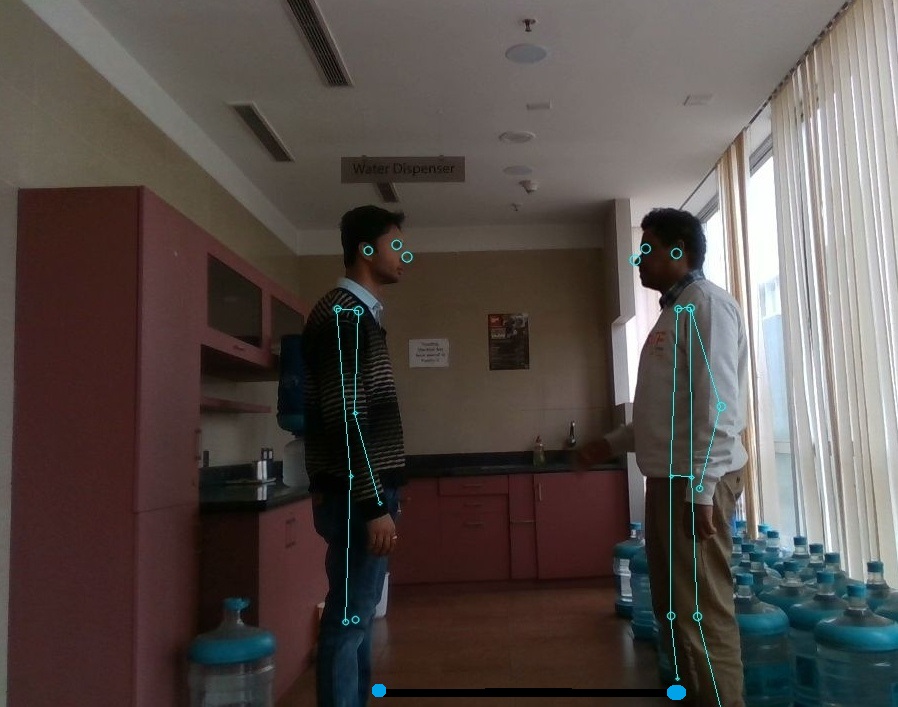}\label{fig:face0}}
	\caption{Triangle, L-shaped, Side-by-side and Face-to-face formations from -90$^\circ$ approach angle with black marks in the floor to define the formation.}
	\label{fig:general_formation}
	\vspace{-3.2mm}
\end{figure}

\textbf{Approach}. In this work, we develop a machine learning based method to detect social groups of people (Fig.~\ref{fig:pipeline}). We use a state-of-the-art deep learning model to detect some key points in the skeleton of the body of all the persons in the scene~\cite{papandreou2017towards, papandreou2018personlab}. Based on these key points, we devise classifiers that take into account the confidence value of each of the key points for training. We create the data-set using a camera mounted on a robot for four real-life formations -- face-to-face, side-by-side, L-shaped, and triangular (Fig.~\ref{fig:general_formation}) from various angles and distances.

We tackle the problem of detecting social groups and the probable outliers in a scene. The forming of clusters of people into a group and then find the group of interest so that it can be further considered for F-formation detection. Then, we develop a robust real-time F-formation detection system that works atop the pre-processing model of avoiding outliers suiting it for robotic applications. Our system is not hampered due to the outliers in a scene. The system also gives the approach angle for the detected F-formation, i.e., the F-formation's orientation w.r.t. the robot or vice versa. This approach angle can be utilized if the robot wants to join the group and find a suitable spot for itself within the group.

\textbf{Contributions:} We have developed an end-to-end system that can detect the formation of social groups (F-formation) from an ego-vision camera in real-time. Our main contributions are listed in the following.
\begin{itemize}
	\item We have developed a robust and highly accurate multi-class SVM based F-formation classifier (Sections~ \ref{sec:2} and \ref{sec:formation}).
	\item To decide the final pose of the robot while joining the group, we also detect the approach angle detection of the F-formation by the robot. This provides the angle of orientation of the group w.r.t. the robot (Sections~\ref{sec:2} and \ref{sec:angle}).
	\item We detect social groups of people in a scene by avoiding outliers using our own CRF probabilistic model. This helps to detect who is part of the same p-space (Sections \ref{subsec:1} and \ref{sec:outlier}).
\end{itemize}

\section{Survey of related works}\label{sec:rel_work}
The problem of detecting social groups and interaction setups is the focus area of this paper. Setti \textit{et al.}~\cite{setti2015f} provides a lucid literature study of the various group detection methods and algorithms with meaningful taxonomies. In this section, we summarize some of the relevant works dealing with similar problems. 

F-formation is used for formally defining a structure in a social gathering where people interact with each other~\cite{kendon1990conducting}. The structures resemble certain geometric shapes like triangle, circle, square, etc. The predominant approaches use rule-based classifiers to detect F-formations~\cite{krishna2017towards,pathi2017estimating}. They use the Haar-cascade classifier to detect faces in an image sequence and then divide the face into four quadrants. The eye placement in the quadrants determines whether the head is oriented left or right. This although doesn't give a very accurate orientational information but serves the purpose. Alternatively, learning-based approaches are also explored in the literature. For example, Hedayati \textit{et al.}~\cite{hedayati2019recognizing} proposed a machine learning model for F-formation detection. The technique, first, deconstructs the image frames to finds all possible pairs of people. Then these pair-wise data with labels are fed into the model to classify if they are a part of any F-formation or not. And, finally, the pair-wise data are reconstructed into the full F-formation data. This method uses the Kinect mounted on a real robot to perceive 3D human poses. Similarly, Aghaei \textit{et al.}~\cite{aghaei2016whom} proposed a model inspired by Spatio-temporal aspects of any interaction among humans. They use two-dimensional time-series data, i.e., distances among participants over time and orientation of the participants over space and time. This sequence of images data is being used to train an LSTM based network. Ego-centric vision is used in the method and real-world scenarios are tested successfully with good accuracy. Alletto \textit{at al.}~\cite{alletto2014ego} proposed yet another data-driven technique using Structural SVM. The article employs correlation-based clustering to assign a pair of people into social interactions or groups. The orientation and distance parameters are used to decide the affinity of two people in a group or interaction. The paper also attends to the dynamics of interacting groups and the varying meanings of distances and orientations among people in different social setups by introducing a Structural SVM to learn the weight vectors of correlation clustering. The method has been tested with ego-centric datasets although not implemented in a real robot. The paper says that it achieves real-time response in detecting interactions. Another contribution gives a graph-cuts minimization based algorithm to cluster participants of a social setup into groups \cite{setti2015f}. The graph-cut finds the o-space in the group of people whose transactional segments (the area in front of any participant/robot/human where the sense of hearing and viewing is maximum) overlap. The transactional segments are identified by orientation of head, shoulder, or feet. They have reported good results for most of the interaction scenarios in real-time. 

Alternatively, Vazquez \textit{et al.}~\cite{vazquez2015parallel} talks about a new exo-centric method to track people's lower body orientation (dynamic Bayesian Network) and head pose detection based on their position and objects of interest in their interaction vicinity, after that group detection (Hough voting scheme) takes place for the participating people in the scene. They also use the concept of soft assignment of participants to o-spaces which in turn allows much faster recovery from errors in group detection of low body orientation trackers. The visual focus of attention is an important cue in understanding social interaction quality and estimation. Bazzani \textit{et al.}~\cite{bazzani2013social} focus on finding the Visual Focus of Attention (VFOA) of a person in 3D.
The paper defines Subjective View Frustum (SVF) for the 3D representation and an Inter-Relation Pattern Matrix for capturing interactions between people in the scene. The paper \cite{mead2013automated} uses a social scientific approach to analyze human proxemics and behaviors in group interaction. The three main parameters being explored are - individual, physical, and psychological. HMMs are trained using two different feature representations: physical and psychological to identify Spatio-temporal behaviors. The significance of these behaviors is marked as the starting and ending of social interaction. The use of two different feature representations helps in a more robust proxemics behavior representation for a robot to be deployed in social setups for interactions. In a recent work~\cite{hedayati2020comparing}, the authors make an interesting study on the formation of groups for interactions where a person is present in the group through a telepresence robot in one case and physically present in another. They found differences in group formation comparing these two cases. However, they established that the positions of each of the participants can be estimated based on the number of participants in the group and the group moderator's position. 

Inspecting the related works we see that the main limitation is the use of ML/DL based methods due to the unavailability of appropriate datasets. Rule-based methods fail in dynamic environments and occlusions. Some methods consider only head pose and orientation which makes the system less robust to dynamism. We try to attend to these limitations in our work. 

\begin{figure*}
	\centering
	t    \includegraphics[width=0.75\linewidth]{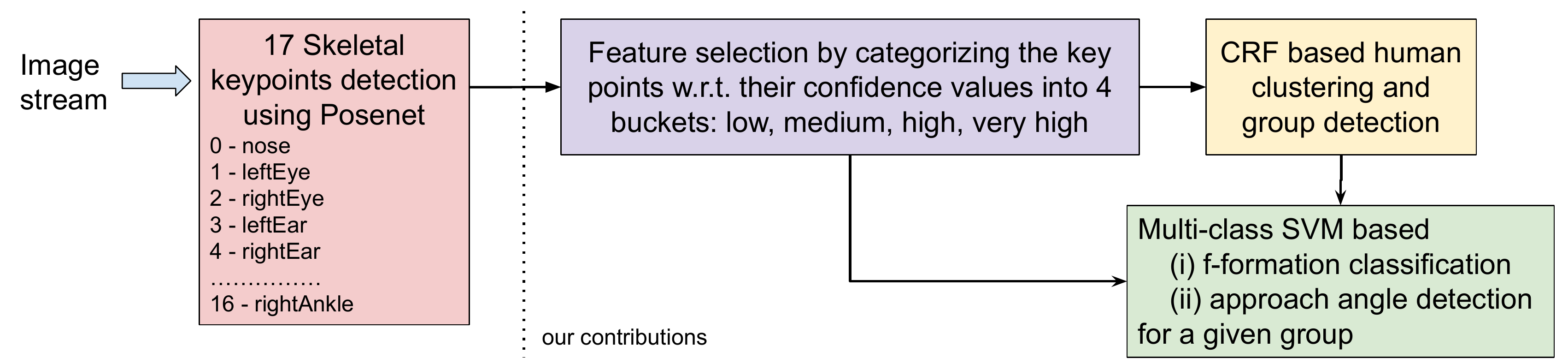}
	\caption{The overview of the pipeline of real-time F-formation classification and approach angle detection.}
	\label{fig:block}
\end{figure*}

\section{Our System: Overview and Design}
\label{sec:prob}

An overview of our system is shown in Fig.~\ref{fig:block}. It consists of two main sub-systems -- a) the group detection and human clustering avoiding outliers, and b) F-formation detection with the type (classification) and approach angle prediction. The first component clusters humans in a scene into socially interacting groups based on some spatial and orientation features. It also detects outliers in the scene to exclude them from groups. The second component finds the F-formation type for each of the detected groups from the previous component. Moreover, it also detects the orientation of the F-formation w.r.t. to the viewing camera (the robot) or the approach angle. This approach angle helps in estimating the optimal pose of the robot to join the group of people. In the following, we discuss all the components of this pipeline in detail.

\subsection{Skeletal key point detection}
\label{subsec:0}
To detect groups in a scene, first, we need to detect all the people in a scene and their corresponding body orientation. Then we need to detect the relationship of each of the persons with another one in the scene. There exist several deep learning-based models that can detect multi-person body skeleton and provide a number of key points corresponding to each person. In this work, we use Posenet\footnote{https://github.com/tensorflow/tfjs-models/tree/master/posenet} to find multi-person skeletal points. As Posenet is a light-weight and highly accurate model, we use it to achieve real-time F-formation detection. Posenet typically yields 17 key points (with co-ordinates) of a human skeleton with some confidence ranging from 0 to 1. The skeleton includes the following human body points- 
\textit{nose, leftEye, rightEye, leftEar, rightEar, leftShoulder, rightShoulder, leftElbow, rightElbow, leftWrist, rightWrist, leftHip, rightHip, leftKnee, rightKnee, leftAnkle, and rightAnkle}. Then these key-points are utilized to find the individual and relative orientation of the people, which leads to F-formation detection.

\subsection{Feature selection}
\label{subsec:01}
As mentioned earlier, Posenet processes an image frame and provides the skeletal key points of each person in the scene. It associates a confidence value with each key point prediction. We assign four labels to each of these key points based on their confidence value -- 
\textit{\{{\small Low: O - 0.25, Medium: O.25 - 0.5, High: O.5 - 0.75, Very high: 0.75 - 1}\}}. The list of key point coordinates along with their labels are used as the feature vector for our classifiers.

\subsection{Groups and outlier detection}
\label{subsec:1}
The problem of group identification and detection in a populated scene is non-trivial. As there can be different types of F-formation with varying number of people along with different body orientation, it poses a challenge who are part of the group. Also, there can be some person standing nearby who is not part of any formation/group; thus should also not be considered for F-formation detection (outliers).

We use a conditional random field (CRF) based probabilistic model to formulate feature functions to predict whether the detected people in a scene are a part of a group or not (see Fig.~\ref{fig:f_formation_spaces} for outliers in a scene). As mentioned earlier, we use the label marking based on the confidence value to create the feature vector. The confidence value of a key point is dependent on the orientation of the person of which the skeleton is being detected by Posenet. Moreover, the location of the corresponding key points of two people can vary significantly based on their closeness or distance between them in the scene. We select such features for our CRF to detect social groups and outliers in a scene.

Given a left-to-right sequence of 2d-poses $p_1, p_2, \dots, p_n$, of the people present in an image, we use a CRF to predict a sequence of group membership label, $g_1, g_2, \dots ,g_n$ , where $g_i \in {G, O}$, where $G$ and $O$ denotes a person in a group and an outlier, respectively. We use the following CRF for this prediction,
\vspace{-0.3cm}
\begin{align*}
	P(g_{1:n}|p_{1:n})= \alpha \exp \bigg \{\sum_{i=0}^n \sum_{j=0}^k \\ \lambda_j f_j(p_{i-1},p_i,p_{i+1},g_{i-1},g_i)  \bigg \},
\end{align*}
where, each node of the CRF uses observation feature functions $f_j$ defined over the current pose and its left and right neighbors, $k$ is the number of such features functions, including a transition feature and $\alpha$ is the normalization factor. 

\begin{figure*}
	\centering
	\subfloat[face-to-face]{\includegraphics[width=0.22\textwidth]{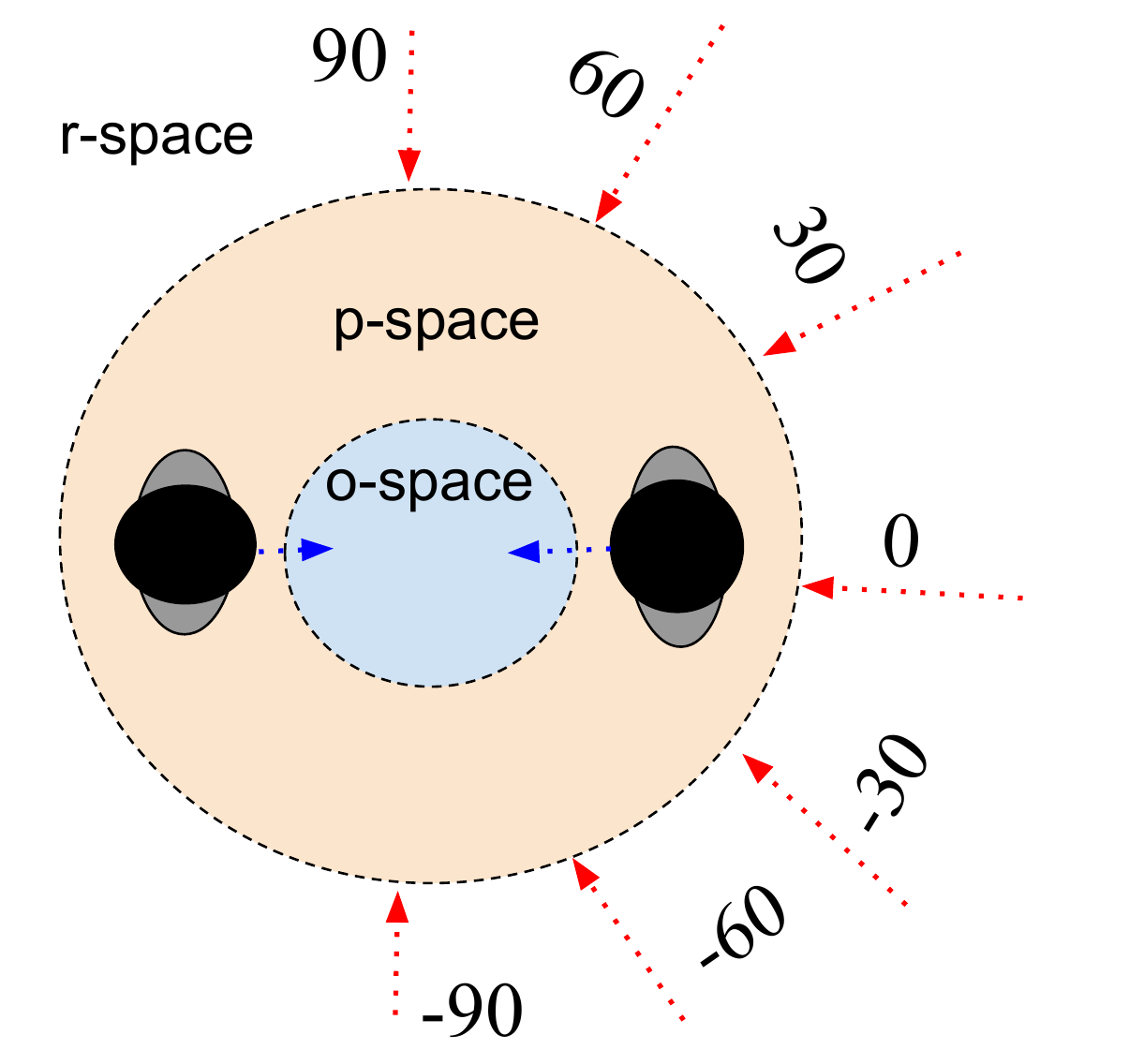}\label{fig:approach1}}
	\subfloat[side-by-side]{\includegraphics[width=0.22\textwidth]{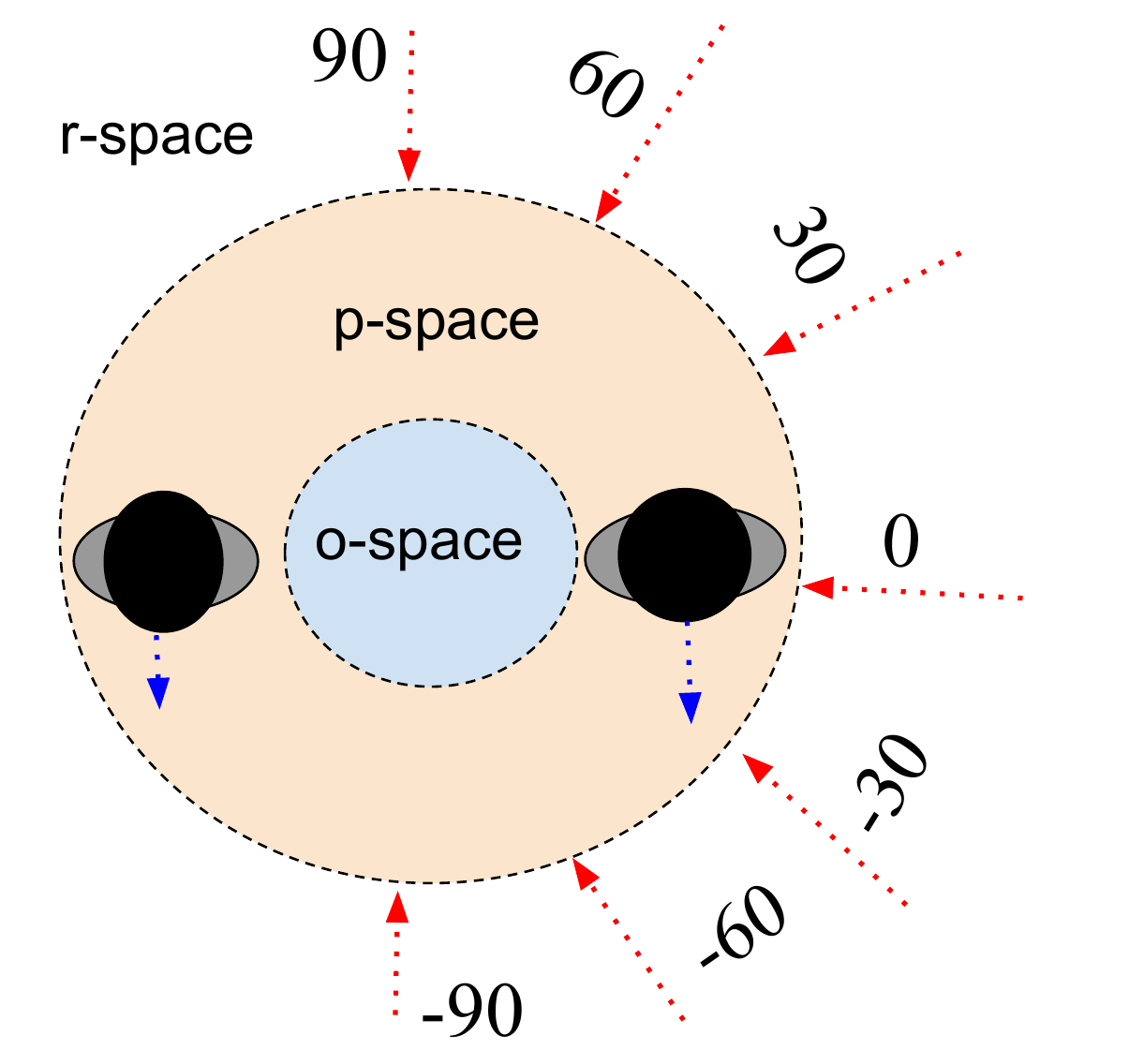}\label{fig:approach2}}
	\subfloat[L-shaped]{\includegraphics[width=0.21\textwidth]{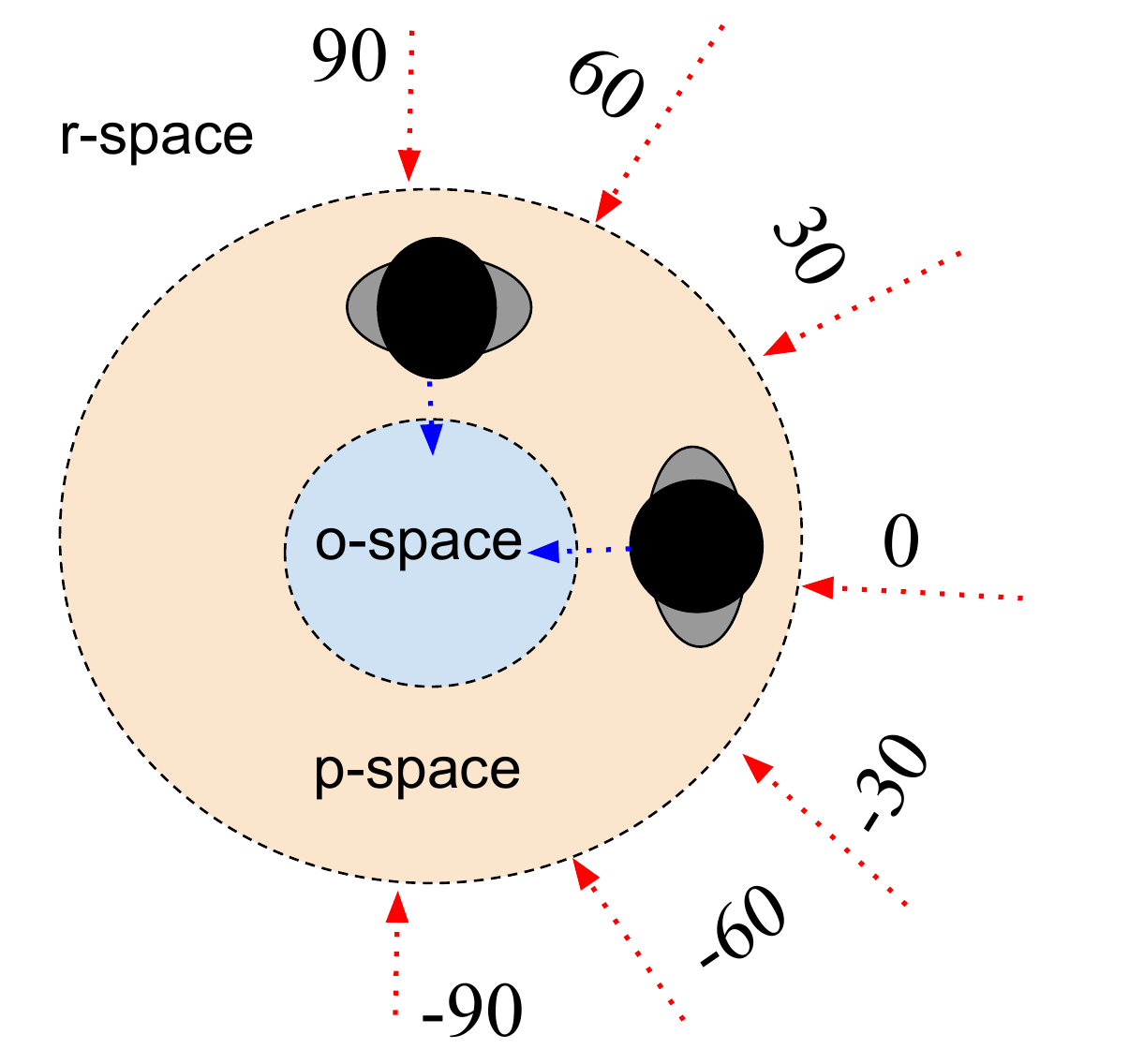}\label{fig:approach3}}
	\subfloat[triangle]{\includegraphics[width=0.22\textwidth]{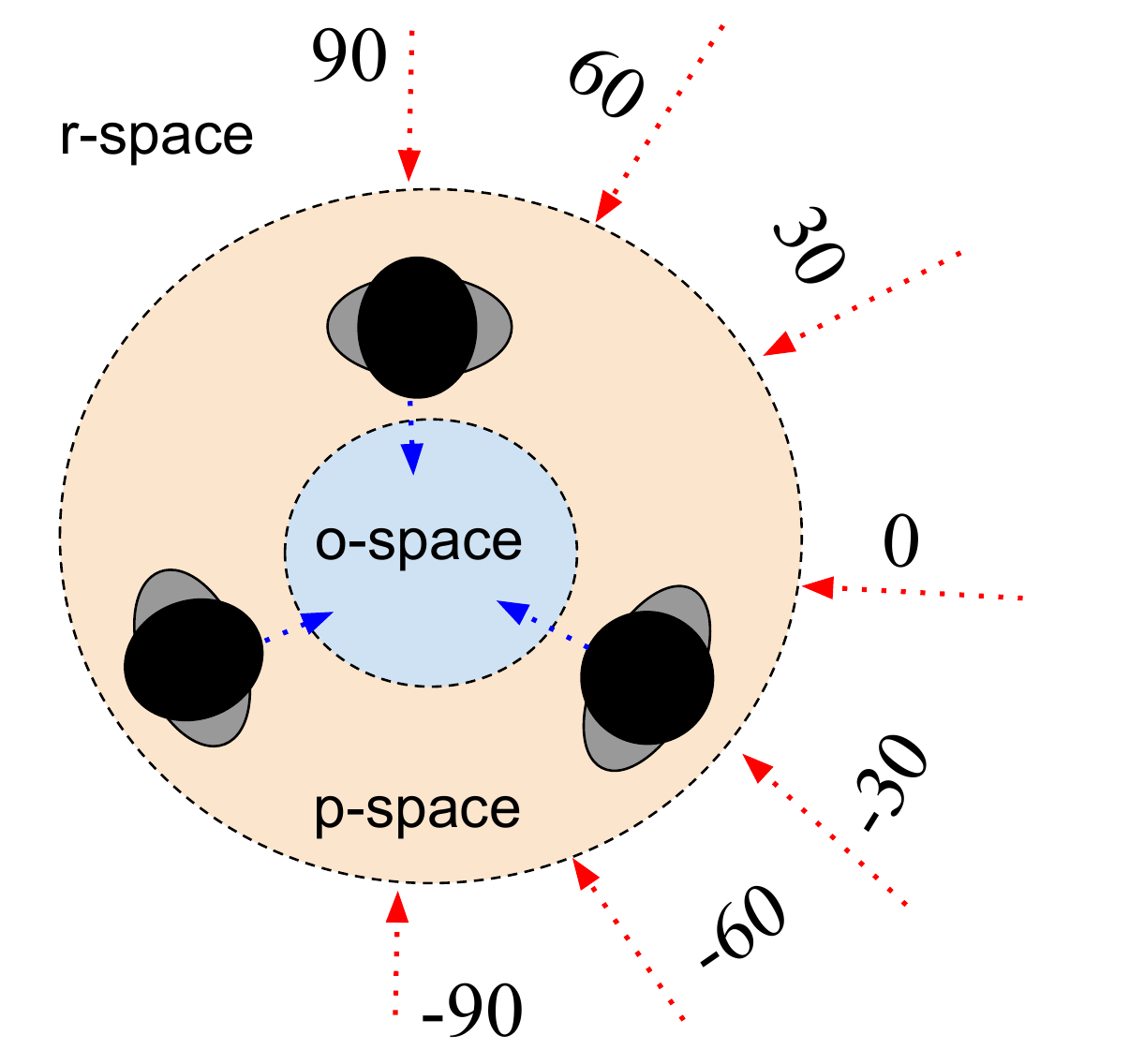}\label{fig:approach4}}

	\caption{Four different F-formations with the considered approach angles layout map (red dotted arrows signify the approach/joining angles of a robot and blue dotted arrows are the orientations of the people in group interaction/formation).}
	\vspace*{-5mm}
	\label{fig:approach}
\end{figure*}
%\vspace*{-5mm}

\subsection{F-formation detection with approach/robot joining angle} 
\label{sec:2}
We use a multi-class Support Vector Machine (SVM) with a Gaussian Radial Bias Function (RBF) kernel, both for the prediction of the F-formation and approach angle. This kernel minimizes the weighted squared Euclidean distance between the feature vectors of two arbitrary samples $X_i, X_j$, i.e.,
$$ K(X_i, X_j) = \exp - \gamma || X_i - X_j || ^2 ,$$
where the free parameter $\gamma$ is learnt from the training data.

Given a set of person-poses after the filtering by the group and outlier detection module, we use the SVM classifier for predicting the F-formation, and based on this prediction, we predict the approximate approach angle. We use the same feature vector for both the prediction, except for approach angle prediction, we use the output of the F-formation classifier as an additional feature and also filter it based on the set of people who belong to the same group. We also show an approach to jointly predict the F-formation and the approach angle in a single classifier and show its efficacy against a baseline approach in Section~\ref{sec:eval}.

We consider the angles -- 0\degree, 30\degree, 60\degree, 90\degree, -30\degree, -60\degree\     and -90\degree\ from which the formation is being viewed by the robot. This prediction reveals the orientation of the formation, thus the joining angle for the robot can be estimated. Fig.~\ref{fig:approach} shows an approach/joining angle layout map that we consider for our leaning models. We consider a complete set of angles for a robot that covers the major part of the formations. We consider -90\degree\ to be the angle/direction which is the optimal angle for a robot to join the formation (in case of L-shaped and triangle) and -90\degree\ as well as 90\degree\ in case of face-to-face and side-by-side depending on the situation. So, if the robot detects the formation from any other angle than -90\degree\ and 90\degree\ (as per the formation), it can move to either left or right direction towards a -90\degree\  or 90\degree\ angle (as per the formation and situation) and then it can move forward to join the group using some distance estimation method such as \cite{pathi2019novel} and stopping criteria as elaborated in the literature survey \cite{ruijten2017stopping}.  

The Face-to-face formation (Fig.~\ref{fig:approach1}), is symmetrical, and therefore the same joining angle can be considered for the -90\degree, -60\degree, -30\degree, and 0\degree\ angles and their symmetrical angles. As for the side-by-side formation (Fig.~\ref{fig:approach2}), both the below 0\degree\ and above 0\degree\ angles are non-symmetric. Here also the remaining semi-circle is irrelevant as its symmetric. For Fig.~\ref{fig:approach3}, which shows the L-shaped formation, the entire set of angles are relevant due to its non-symmetric nature. For the triangle formation in Fig.~\ref{fig:approach4} the same argument holds tight.

\section{Experiments and Evaluation}
\label{sec:eval}
In this section, we evaluate our system on two aspects -- the CRF probabilistic model for group and outlier detection and the multi-class SVM model with a radial bias function (RBF) kernel for F-formation detection and approach angle prediction. We only compared the accuracy of F-formation detection with a rule-based state-of-the-art system as proposed by Pathi \textit{et al.}~\cite{pathi2017estimating}. Since multiple group detection and approach angle detection is not done by any existing literature, we could not compare it with any.

\subsection{Datasets and Setups}
\label{sec:data}
For training and testing our models, we have used two sets of data -- the EGO-GROUP\footnote{http://imagelab.ing.unimore.it/files/EGO-GROUP.zip} \cite{alletto2014ego} dataset and a locally collected dataset. The EGO-GROUP dataset contains a total of 2900 frames annotated with groups. In different frames, they have used 19 different human subjects. In has 4 different scenarios to make it more challenging for detection. The dataset considers a laboratory setting, a coffee break setting, a festive scenario, and an outdoor scenario. All these scenarios have different levels of background clutter, lighting conditions, and orientation settings.   

We have also created a local dataset of four common F-formations -- face-to-face, L-shaped, side-by-side, and triangle. For each of these formations, we collected the data using 4 different subjects from 7 different angles -- 0\degree, 30\degree, 60\degree, 90\degree, -30\degree, -60\degree\  and -90\degree\ as shown in Fig.~\ref{fig:approach}. Moreover, for every formation and every angle, we have collected data from different distances from the center of the o-space ranging from 2 meters to 5 meters. Further, we considered different lighting conditions, backgrounds, partial occlusion, and cluttered background while collecting the data. The dataset consists of 4692 image frames in total, which are annotated with the four formations and the approach angles. We collected the images from a video stream using a RealSense\footnote{https://www.intel.in/content/www/in/en/architecture-and-technology/realsense-overview.html} camera mounted on a Double2\footnote{https://www.doublerobotics.com/double2.html} robot. We make an 80\%-20\%  split (for each formation) of the dataset for training and testing correspondingly. The experiments have been conducted with a single-core CPU with 8GB RAM (even lower memory is fine) without any graphics processing units (GPU). Thus, our system can be run on any embedded system or robotic hardware.  

\subsection{Group membership classification results}
\label{sec:outlier}
The CRF model for group and outlier detection is trained and tested using a subset of the EGO-GROUP dataset which contains 969 image frames, with an avg. of 3.6 persons per frame, where we used an 80-20\% split for training and testing. We selected only those frames that contain the four F-formations and outliers to make the subset. The system displays an average accuracy of \textbf{91\%} in detecting the outlier, who are not part of the interacting group. Table~\ref{tab:group} shows the evaluation metrics for the two membership classes. By analyzing the failure cases, we find the classifier makes errors in some frames where the outlier is in close proximity to the group and also has a very similar orientation. However, this can be mitigated by introducing temporal information in the classification. Overall, the detection of social groups before the prediction of F-formation should contribute to the robustness of the system. 
\begin{table}[t]
	\caption{Results for the prediction of group membership sequence given the sequence of 2D poses.}
	\centering
	\begin{tabular}{|c|c|c|c|}
		\hline
		\textbf{Group membership} & \textbf{Precision} & \textbf{Recall} &\textbf{F1} \\ \hline
		Group       &0.92      &0.92      &0.92 \\
		Outlier       &0.89      &0.89      &0.89 \\
		\hline
		\textbf{ Avg.}  &\textbf{0.91}      &\textbf{0.91}      &\textbf{0.91} \\
		\hline
	\end{tabular}
	
	\label{tab:group}
\end{table}
\begin{savenotes}
	
	\begin{table}[t]
		\centering
		\caption{F-formation prediction results for our method (learning based) and rule based method.}
		\begin{tabular}{|c|c|c|c|c|}
			\hline
			\textbf{F-formation} & \textbf{Precision} & \textbf{Recall} & \textbf{F1} & \textbf{Rule based method\footnote{For multi-class classification, we use weighted avg. F1 score and accuracy interchangeably.}} \\ \hline
			face-to-face       &0.99      &0.99      &0.99 & 0.68\\
			side-by-side       &0.95      &0.99      &0.97 & 0.94\\
			L-shaped       &1.00      &0.96      &0.98 & 0.49 \\
			triangle       &1.00      &1.00      &1.00 & 0.63 \\ \hline
			\color{blue}{ \textbf{ Avg.}}  & \color{blue}{\textbf{0.98}}      & \color{blue}{\textbf{0.98}}      & \color{blue}{\textbf{0.98}}  & \color{blue}{\textbf{0.69}}\\
			\hline 
		\end{tabular}
		
		\label{tab:F-formation}
	\end{table}
\end{savenotes}
\begin{table}[t]
	\centering
	\caption{Approximate approach angle prediction results. }
	\begin{tabular}{|c|c|c|c|}
		\hline
		\textbf{Approximate angle} & \textbf{Precision} &\textbf{Recall} & \textbf{F1} \\ \hline
		-90$^\circ$       &0.93      &0.98      &0.95       \\
		-60$^\circ$       &0.98      &0.95      &0.96       \\
		-30$^\circ$       &0.98      &0.99      &0.98       \\
		0$^\circ$       &0.94      &0.99      &0.97\\    30$^\circ$       &0.97      &0.89      &0.93       \\
		60$^\circ$       &0.93      &0.94      &0.94       \\
		90$^\circ$       &0.97      &0.95      &0.96       \\
		\hline
		\textbf{Avg.}  & \textbf{0.96}      & \textbf{0.96}      & \textbf{0.96} 
		\\
		\hline
	\end{tabular}
	
	\label{tab:my_label}
	\vspace{-2.2mm}
\end{table}
\vspace{-1mm}

\begin{table}[t]
	\caption{Accuracy of the joint prediction of F-formation and approach angle.}
	\begin{tabular}{|p{2cm}|p{3.1cm}|p{2.2cm}|}
		\hline
		\textbf{F-formation and angle} &\textbf{Our learning based method (joint prediction accuracy for F-formation and approach angle)} & \textbf{State-of-the-art (prediction accuracy for only F-formation)}  \\ \hline\hline
		face-to-face -90    & 80\% &  54\%    \\
		face-to-face -60  & 86\%  & 43\%   \\
		face-to-face -30      & 96\% & 45\%     \\
		\cellcolor{pink}face-to-face 0      & \cellcolor{pink}100\% & \cellcolor{pink}0\%     \\ 
		face-to-face 30 & 98\% & 22\%    \\
		face-to-face 60 & 95\% &  3\%   \\
		\cellcolor{cyan}face-to-face 90  & \cellcolor{cyan}70\%  & \cellcolor{cyan}82\%   \\ \hline\hline
		\cellcolor{cyan}side-by-side -90       & \cellcolor{cyan}91\% & \cellcolor{cyan}94\%    \\
		\cellcolor{cyan}side-by-side -60      & \cellcolor{cyan}82\%  & \cellcolor{cyan}96\%    \\
		side-by-side -30      & 88\%  & 86\%   \\
		\cellcolor{pink}side-by-side 0      & \cellcolor{pink}96\%  & \cellcolor{pink}0\%   \\
		side-by-side 30   & 83\%  & 21\%  \\
		side-by-side 60      & 88\% & 2\%    \\
		side-by-side 90     & 97\%  & 2\%  \\  \hline\hline
		L-shaped -90             & 99\% &   49\% \\
		L-shaped -60              & 98\%  & 29\%   \\
		L-shaped -30             & 82\% & 6\%    \\
		L-shaped 0             & 90\%  & 25\%  \\
		L-shaped 30             & 81\%  &  10\% \\
		L-shaped 60            & 93\%  & 30\%  \\
		L-shaped 90           & 83\%  &  1\% \\  \hline\hline
		triangle -90             & 97\%  & 48\%  \\
		triangle -60             & 100\% & 56\%   \\
		\cellcolor{pink}triangle -30            & \cellcolor{pink}90\%   & \cellcolor{pink}0\% \\
		triangle 0             & 100\%  &  56\% \\
		\cellcolor{pink}triangle 30          & \cellcolor{pink}100\%  & \cellcolor{pink}0\%  \\
		triangle 60           & 99\%  &  91\% \\
		triangle 90           & 97\%  &  78\% \\ \hline\hline
		\textbf{Avg.}       &\textbf{92\%} & \textbf{37\%}   \\ 
		\hline
	\end{tabular}
	
	\label{tab:accuracy}
	\vspace{-3.2mm}
\end{table}

\subsection{F-formation classification results}
\label{sec:formation}
As mentioned earlier, we use the encoded skeletal key points vector to train a multi-class SVM classifier for F-formation detection. We compare the accuracy of our system with the state-of-the-art (rule-based) method in \cite{pathi2017estimating}. Table \ref{tab:F-formation} summarizes the prediction accuracy of these two methods. Since the state-of-the-art system is originally tested for only a limited approach angle, i.e., -90\degree\  and 90\degree\ for face-to-face, only -90\degree\ for side-by-side, and only -90\degree\ for L-shaped and finally -90\degree\ and +90\degree\ for triangle, this table represents the accuracy for the subset of data with these approach angles only. It is evident that our system improves the accuracy significantly (\textbf{98\%} as compared to \textbf{69\%}) and achieves a near-perfect accuracy for all the formations with these approach angles. The main reason is that the rule-based method considers only head pose orientation features such as eyes and ears for writing the rules. The head pose/orientation detection is also limited to left, right, and front direction prediction for the rule-based, hence limiting its formation detection capability to a great extent. However, in our learning-based SVM method we consider full body orientation features as given by skeletal key points (e.g., Posenet DL model) and not limited to a fixed number of orientations. This also contributes to its spike in accuracy compared to the other method. Also, the rule-based method doesn't take into account partial occlusion and so it fails if the head and its features are not visible partially whereas our method overcomes that situation through rigorous learning.      

\subsection{Joint prediction of F-formation approach/joining angle}
\label{sec:angle}
The approach angle detection is necessary to understand the orientation of the F-formation with respect to the detecting camera/robot. This facilitates efficient navigation of the robot/device to join the formation at an appropriate spot (as discussed in Section \ref{sec:2}). Table \ref{tab:my_label} reports the result of approach angle prediction (Precision, Recall, and F1-score) for any F-formation using our learning-based method. The results show an acceptable accuracy of \textbf{96\%} for approach angle prediction alone. The problem of approach angle prediction in F-formation is novel to the best of our knowledge and has not been attended so far. The rule-based method in \cite{pathi2017estimating} has not considered approach angle prediction along with F-formation using their rule-based classifier as its head pose/orientation detection rules displays limited capability in detecting formations. However, we have used similar rules to implement a rule-based method. So, we compare these two methods among themselves in Table \ref{tab:accuracy}. The table contains the combined results for F-formation and approach angle prediction. As we can see clearly that our learning-based method has outperformed our rule-based method (similar to rule-based method \cite{pathi2017estimating}) with a high margin. A difference of \textbf{55\%} in overall accuracy establishes the efficacy of our learning-based method over rule-based counterpart.

\begin{figure}
	\centering
	\subfloat[side-by-side 0]{\includegraphics[width=0.22\textwidth]{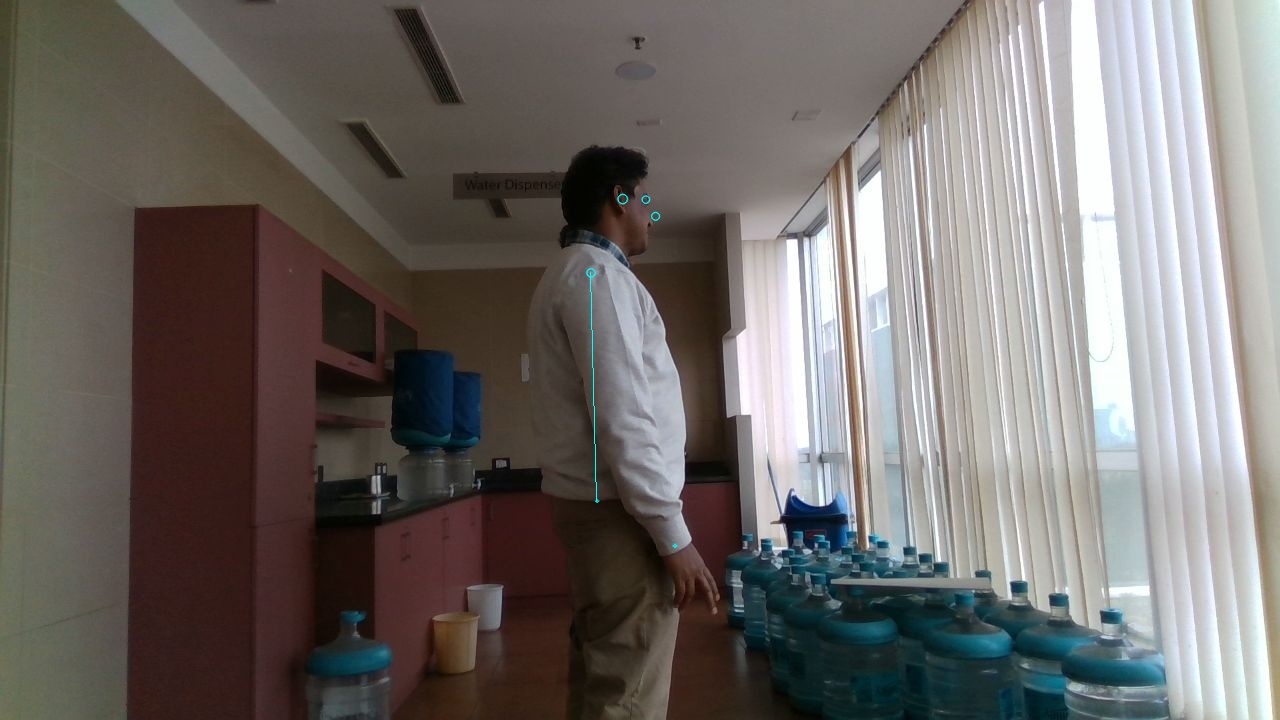}\label{fig:side}}
	\subfloat[face-to-face 0]{\includegraphics[width=0.22\textwidth]{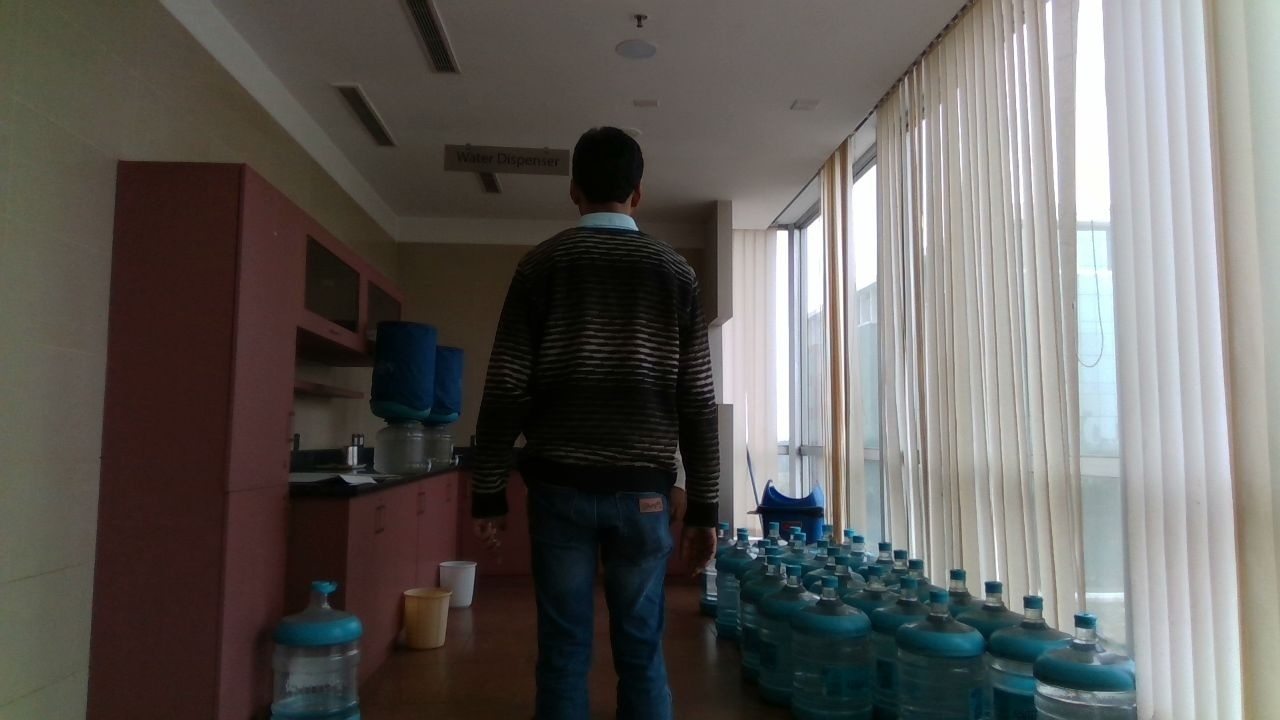}\label{fig:face}}\\
	\subfloat[triangle -30]{\includegraphics[width=0.22\textwidth]{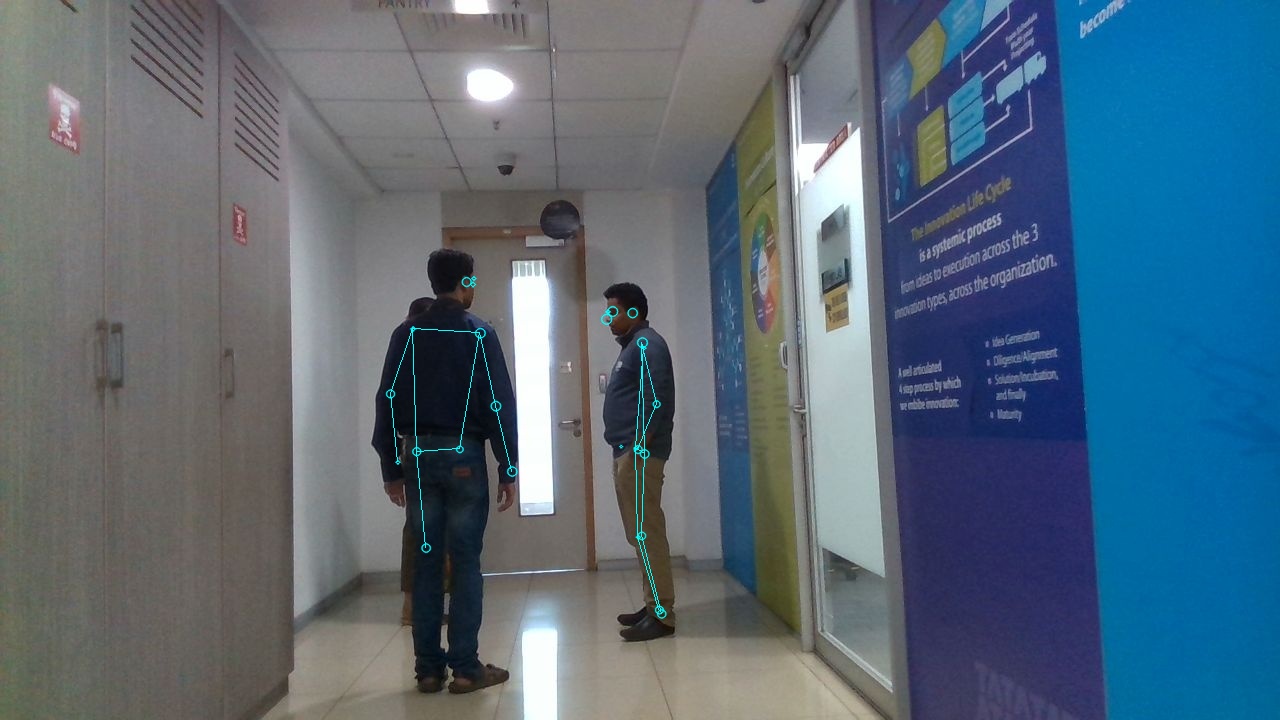}\label{fig:triangle1}}
	\subfloat[triangle 30]{\includegraphics[width=0.22\textwidth]{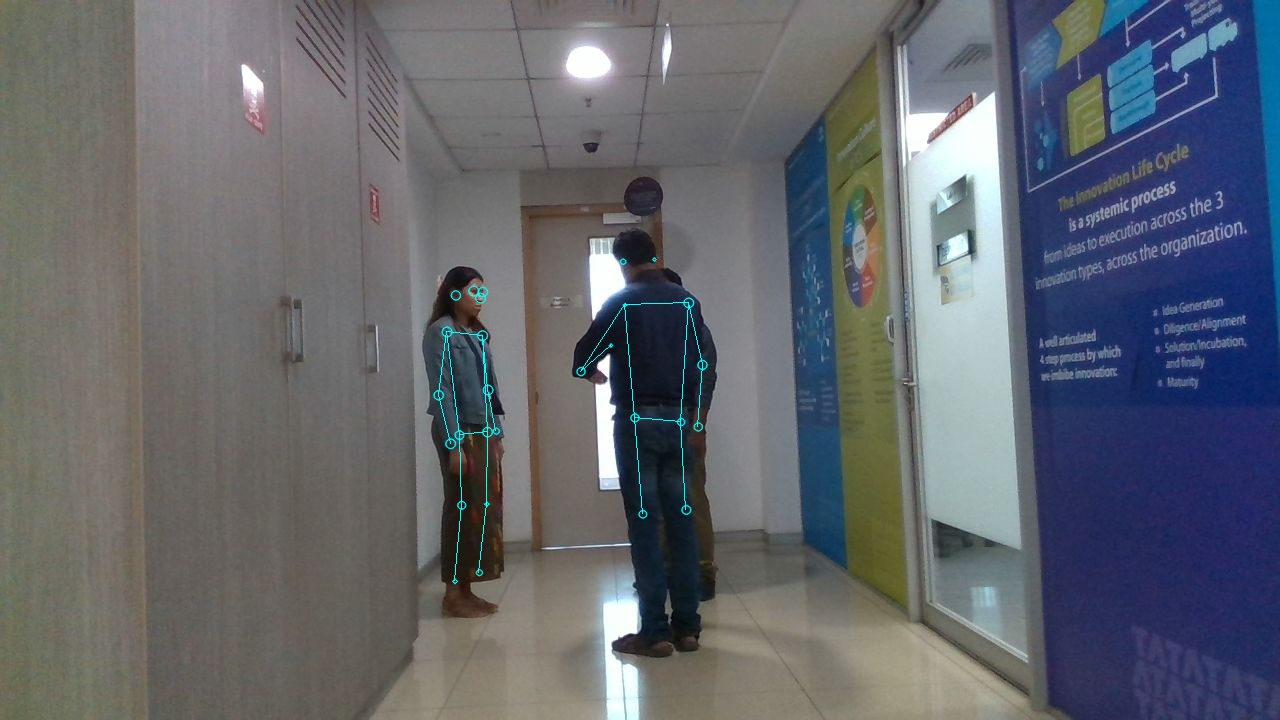}\label{fig:triangle2}}\\

	\caption{Occlusion of one of the member in a Side-by-side, Face-to-face and Triangle formations.}
	\label{fig:triangle_occlusion}
	\vspace{-2.2mm}
\end{figure}
%\vspace{-2mm}
\begin{figure}
	\centering
	\subfloat[L-shaped -30]{\includegraphics[width=0.22\textwidth]{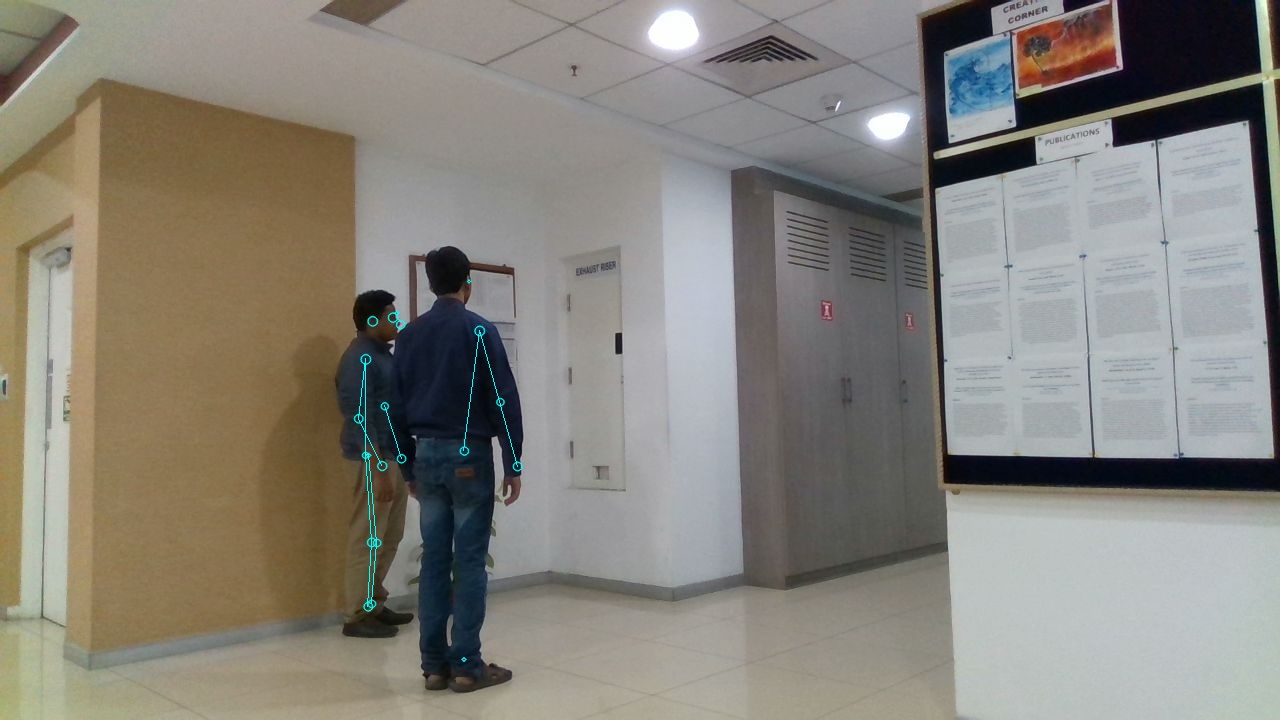}\label{fig:lshaped1}}
	\subfloat[L-shaped 90]{\includegraphics[width=0.22\textwidth]{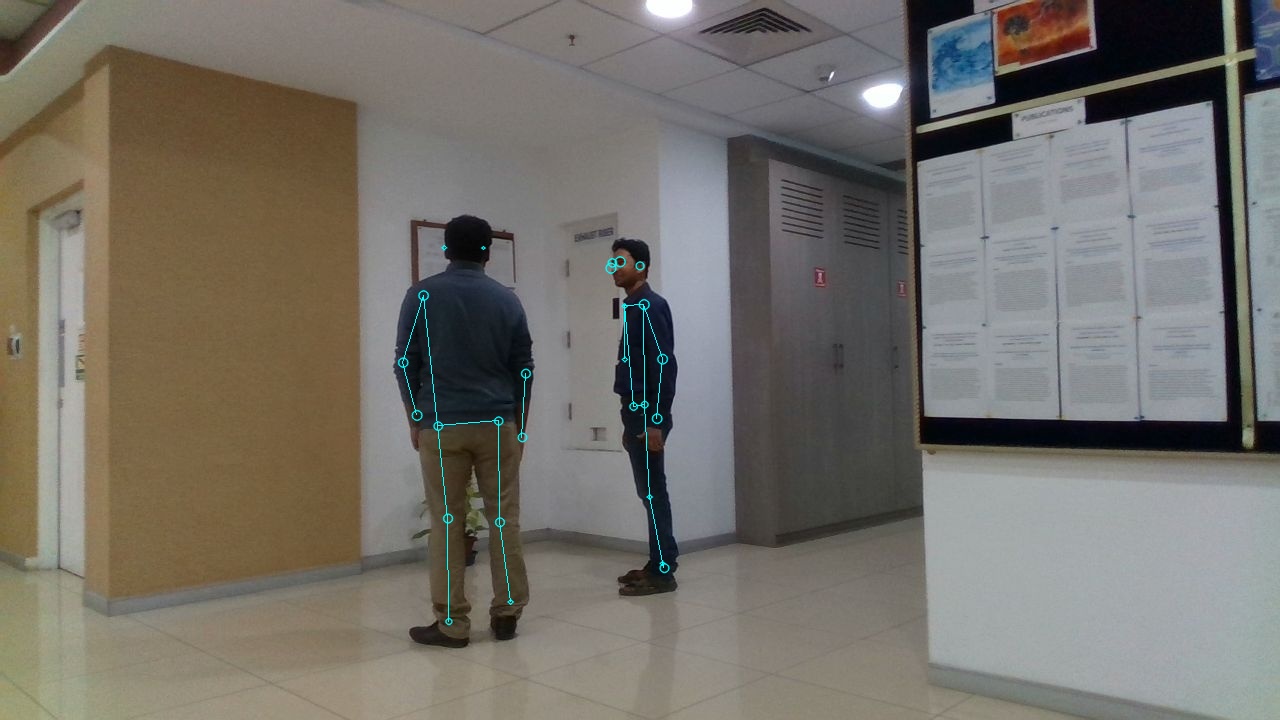}\label{fig:lshaped2}}
	\caption{L-shaped formations.}
	\label{fig:lshaped_occlusion}
	\vspace{-2.2mm}
\end{figure}

Table \ref{tab:accuracy} has some interesting results if we look at the individual rows. The rows in \textit{pink} color gives us those formation with approach angle where rule-based method failed completely. For side-by-side 0 and face-to-face 0 (see Fig. \ref{fig:side} and \ref{fig:face}), the person nearer to the camera occludes the other person within the formation highly. So, the rule-based method, where only head pose orientation along with eyes and ears position is considered, fails in this case. But in our learning model, since we use the entire human body pose features for prediction (not simply relying on the head pose, orientation, and eye/ear positions) it gives excellent results. The triangle -30 and triangle 30 have similar occlusions of one of the members of the formation, hence the failure in rule-based method (see Fig. \ref{fig:triangle1} and \ref{fig:triangle2}). However, in these two cases, miss-classifications have taken place more than 50\% of times, predicting them as face-to-face, side-by-side, or L-shaped in some cases due to occlusion of one member. This is clearly shown in the four cases of Fig.~\ref{fig:triangle_occlusion}.
The other cases with very low accuracy have similar reasons for occlusions and/or miss-classifications in the rule-based method. Fig. \ref{fig:lshaped1} and \ref{fig:lshaped2} show two cases of L-shaped formation where miss-classifications have taken place to a great extent predicting them as side-by-side or face-to-face.  

The rows in \textit{blue} (Table \ref{tab:accuracy}) color gives us the cases where the rule-based method outperforms the learning-based method. These are roughly the cases where the rules as stated in \cite{pathi2017estimating} best suit the conditions. These angles of viewing the F-formations are the best-case scenarios for the rule-based classifier. The other such cases where the rule-based method has quite a high accuracy are pertaining to the same fact. 
There is a  decline of 6\% in the accuracy of our learning-based method while considering approach angle is due to the lesser number of training data samples when the dataset is divided among various approach angles for the different formations.

% Covid 19 social distancing and self adjusting if a member of the group leaves or new member joins.

\definecolor{aqua}{rgb}{0.0, 1.0, 1.0}
\definecolor{aquamarine}{rgb}{0.5, 1.0, 0.83}
\definecolor{ao}{rgb}{0.0, 0.0, 1.0}
\definecolor{blue}{rgb}{0.0, 0.0, 1.0}
\definecolor{blue(ryb)}{rgb}{0.01, 0.28, 1.0}
\definecolor{bisque}{rgb}{1.0, 0.89, 0.77} 	
\definecolor{bananamania}{rgb}{0.98, 0.91, 0.71}
\definecolor{bananayellow}{rgb}{1.0, 0.88, 0.21}
\definecolor{arylideyellow}{rgb}{0.91, 0.84, 0.42}
\definecolor{apricot}{rgb}{0.98, 0.81, 0.69}
\definecolor{aureolin}{rgb}{0.99, 0.93, 0.0}
\definecolor{awesome}{rgb}{1.0, 0.13, 0.32}
\definecolor{bittersweet}{rgb}{1.0, 0.44, 0.37}
\definecolor{dandelion}{rgb}{0.94, 0.88, 0.19}
\definecolor{buff}{rgb}{0.94, 0.86, 0.51}
\definecolor{cyan(process)}{rgb}{0.0, 0.72, 0.92}
\definecolor{cottoncandy}{rgb}{1.0, 0.74, 0.85}
\definecolor{corn}{rgb}{0.98, 0.93, 0.36}
\definecolor{flavescent}{rgb}{0.97, 0.91, 0.56}
\definecolor{babyblue}{rgb}{0.54, 0.81, 0.94}
\definecolor{beaublue}{rgb}{0.74, 0.83, 0.9}
\definecolor{blizzardblue}{rgb}{0.67, 0.9, 0.93}
\definecolor{columbiablue}{rgb}{0.61, 0.87, 1.0}

\section{Conclusions and Future works}
\label{sec:conc}
This work presents a novel method to detect the formation of a social group of people, a.k.a., F-formation in real-time in a given scene. We can also detect outliers in the process, i.e., people who are visible but not part of the interacting group. This plays a key role in correct F-formation detection in a real-life crowded environment. Additionally, when a collocated robot plans to join the group it has to detect a pose for itself along with detecting the formation. Thus, we also provide the approach angle for the robot, which can help it to determine the final pose in a socially acceptable manner. Although considerable works are present in this domain, accuracy is limited due to the prevalent rule-driven methods. On the other hand, the existing learning-based approaches lack robustness under varied conditions of light, occlusion, and backgrounds. We have created a dataset catering to such scenarios. Our method of jointly detecting formation orientation and approach angle for a robot is not addressed in any of the existing literature to the best of our knowledge. The results show that our system outperforms the state-of-the-art method significantly in detecting F-formations correctly.

%\subsection{Future works}
An immediate extension of this work can be detecting other F-formations along with the 4 major types discussed in this article. This work also puts forward a novel idea of detecting the orientation of the formation to understand where should a robot navigate to join the group. But, the main limitation here is a human-aware navigation strategy to be stitched with this method so that the robot can navigate to join the formation after detecting the F-formation class and approach angle satisfactorily. Going further a human/user experience study is to be conducted for the entire system of detecting and joining a group by a robot. Additionally, we also intend to extend the system by detection of multiple groups in a scene along with outliers which can facilitate monitoring of new social norms in COVID-19 scenario. %We also intend to enrich our dataset with a much higher number of samples for training in various environmental/human conditions.      

\addtolength{\textheight}{-12cm}   % This command serves to balance the column lengths

%%%%%%%%%%%%%%%%%%%%%%%%%%%%%%%%%%%%%%%%%%%%%%%%%%%%%%%%%%%%%%%%%%%%%%%%%%%%%%%%

%%%%%%%%%%%%%%%%%%%%%%%%%%%%%%%%%%%%%%%%%%%%%%%%%%%%%%%%%%%%%%%%%%%%%%%%%%%%%%%%

%%%%%%%%%%%%%%%%%%%%%%%%%%%%%%%%%%%%%%%%%%%%%%%%%%%%%%%%%%%%%%%%%%%%%%%%%%%%%%%%
%\section*{APPENDIX}
\section*{ACKNOWLEDGMENT}
We would like to thank Mr. Sai Krishna Pathi of Örebro University to provide us the source code of his work that we utilized for comparing performance of our system. We would also like to thank Mr. Sayan Paul and Ms. Marichi Agarwal of TCS Research for helping us in collecting the data.%The preferred spelling of the word ÒacknowledgmentÓ in America is without an ÒeÓ after the ÒgÓ. Avoid the stilted expression, ÒOne of us (R. B. G.) thanks . . .Ó  Instead, try ÒR. B. G. thanksÓ. Put sponsor acknowledgments in the unnumbered footnote on the first page.

%%%%%%%%%%%%%%%%%%%%%%%%%%%%%%%%%%%%%%%%%%%%%%%%%%%%%%%%%%%%%%%%%%%%%%%%%%%%%%%%
\balance

\bibliographystyle{IEEEtran}
\bibliography{root}

\end{document}